%% file: neurips_2020.tex
\documentclass{article}

\PassOptionsToPackage{square, numbers}{natbib}

\usepackage[final]{neurips_2020}

\usepackage[utf8]{inputenc} % allow utf-8 input
\usepackage[T1]{fontenc}    % use 8-bit T1 fonts
\usepackage{hyperref}       % hyperlinks
\usepackage{url}            % simple URL typesetting
\usepackage{booktabs}       % professional-quality tables
\usepackage{amsfonts}       % blackboard math symbols
\usepackage{nicefrac}       % compact symbols for 1/2, etc.
\usepackage{microtype}      % microtypography
\usepackage{graphicx}
\usepackage{subfigure}
\usepackage{chapterbib}
\usepackage{array, makecell}
\usepackage{tikz-network}
\usepackage{multirow}
\usepackage{lipsum}        
\usepackage{xargs}         
\usepackage[normalem]{ulem}
\usepackage{algorithmic}
\usepackage{algorithm}

\newcommand*{\method}{Graph-$Q$-SAT}
\newcommand*{\minisat}{MiniSat}
\newcommand*{\metric}{MRIR}
\newcommand*{\fullmetric}{Median Relative Iteration Reduction (\metric{})}
\newcommand{\gqsattitle}{Can $Q$-Learning with Graph Networks Learn a\\Generalizable Branching Heuristic for a SAT Solver?}

\usepackage{bm}
\renewcommand{\vec}[1]{\bm{#1}}

\title{\gqsattitle}
\author{Vitaly Kurin\thanks{The work was done when the author was a research intern at NVIDIA.} \\
Department of Computer Science \\
University of Oxford \\
Oxford, United Kingdom \\
\texttt{vitaly.kurin@cs.ox.ac.uk} \\
\And
Saad Godil \\
NVIDIA \\
Santa Clara, California \\
United States \\
\texttt{sgodil@nvidia.com}\\
\\
\AND
Shimon Whiteson \\
Department of Computer Science \\
University of Oxford \\
Oxford, United Kingdom \\
\texttt{shimon.whiteson@cs.ox.ac.uk} \\
\And
Bryan Catanzaro \\
NVIDIA \\
Santa Clara, California \\
United States \\
\texttt{bcatanzaro@nvidia.com} 
}

\begin{document}
\maketitle
\input{sections/abstract.tex}%
\input{sections/introduction.tex}
\input{sections/background.tex}%
\input{sections/method.tex}%
\input{sections/experiments.tex}%
\input{sections/related_work.tex}
\input{sections/discussion.tex}%
\input{sections/impact}
\input{sections/acknowledgements}
\bibliographystyle{abbrvnat}
\bibliography{neurips_2020}
\appendix
\include{sections/appendix}

\end{document}

%% file: sections/abstract.tex
\begin{abstract}
We present \method{}, a branching heuristic for a Boolean SAT solver trained with value-based reinforcement learning (RL) using Graph Neural Networks for function approximation. Solvers using \method{} are complete SAT solvers that either provide a satisfying assignment or proof of unsatisfiability, which is required for many SAT applications. The branching heuristics commonly used in SAT solvers make poor decisions during their warm-up period, whereas \method{} is trained to examine the structure of the particular problem instance to make better decisions early in the search. Training \method{} is data efficient and does not require elaborate dataset preparation or feature engineering. We train \method{} using RL interfacing with \minisat{} solver and show that \method{} can reduce the number of iterations required to solve SAT problems by 2-3X. Furthermore, it generalizes to unsatisfiable SAT instances, as well as to problems with 5X more variables than it was trained on. We show that for larger problems, reductions in the number of iterations lead to wall clock time reductions, the ultimate goal when designing heuristics.
We also show positive zero-shot transfer behavior when testing \method{} on a task family different from that  used for training.
While more work is needed to apply \method{} to reduce wall clock time in modern SAT solving settings, it is a compelling proof-of-concept showing that RL equipped with Graph Neural Networks can learn a generalizable branching heuristic for SAT search.
\end{abstract}

%% file: sections/introduction.tex
\section{Introduction}

Boolean satisfiability (SAT) is an important problem for both industry and academia that impacts various fields, including circuit design, computer security, artificial intelligence and automatic theorem proving. %, and combinatorial optimization.
As a result, modern SAT solvers are well crafted, sophisticated, reliable pieces of software that can scale to problems with hundreds of thousands of variables~\citep{ohrimenko2009propagation}.

SAT is known to be NP-complete~\citep{karp1972reducibility}, and most state-of-the-art open-source and commercial solvers rely on multiple \emph{heuristics} to speed up the exhaustive search, which is otherwise intractable.
These heuristics are usually meticulously crafted using expert domain knowledge and are often iteratively refined via trial and error. 
In this paper, we investigate how we can use machine learning to improve upon an existing branching heuristic without leveraging domain expertise.

We present \method{}, a branching heuristic in a Conflict Driven Clause Learning~\citep[CDCL]{marques1999grasp, bayardo1997using} SAT solver trained with value-based reinforcement learning (RL), based on deep $Q$-networks~\citep[DQN]{mnih2015human}.
\method{} uses a graph representation of SAT problems similar to~\citet{selsam2018learning} which provides permutation and variable relabeling invariance.
\method{} uses a Graph Neural Network~\citep[GNN]{gori2005new, battaglia2018relational} as a function approximator to provide generalization as well as support for a dynamic state-action space.
\method{} uses a simple state representation and a binary reward that requires no feature engineering or problem domain knowledge.
\method{} modifies only part of the CDCL based solver, keeping it \emph{complete}, i.e., always yielding a correct solution.

We demonstrate that \method{} outperforms Variable State Independent Decaying Sum~\citep[VSIDS]{moskewicz2001chaff}, the most frequently used CDCL branching heuristic, reducing the number of iterations required to solve SAT problems by 2-3X.
\method{} is trained to examine the structure of the particular problem instance to make better decisions at the beginning of the search, whereas the VSIDS heuristic suffers from poor decisions during the warm-up period.

Our work primarily focuses on the machine learning perspective and thus more work would be required to apply \method{} in industrial-scale SAT settings.
However, \method{} exhibits intriguing properties which might eventually be useful for practical applications.
We show that our method generalizes to problems five times larger than those it was trained on.
We also show that \method{} generalizes across problem types from satisfiable (SAT) to unsatisfiable instances (unSAT).
We show that reducing the number of iterations, in turn, could reduce wall clock time, the ultimate goal when designing heuristics.
We also show positive zero-shot transfer properties of \method{} when the testing task family is different from the training one.
Finally, we show that some of these improvements are achieved even when training is limited to a single SAT problem, demonstrating data efficiency.

%% file: sections/background.tex
\section{Background}

\subsection{SAT problem}

A SAT problem involves finding variable assignments such that a propositional logic formula is satisfied or showing that such an assignment does not exist.
A propositional formula is a Boolean expression, including Boolean variables, ANDs, ORs and negations.
The term literal is used to refer to a variable or its negation.
It is convenient to represent Boolean formulas in conjunctive normal form (CNF), i.e., conjunctions (AND) of clauses, where a clause is a disjunction (OR) of literals.
An example of a CNF is $(x_1 \lor \neg x_2) \land (x_2 \lor \neg x_3)$, where $\land, \lor, \neg$ are AND, OR, and  negation respectively.
This CNF formula has two clauses: $(x_1 \lor \neg x_2)$ and $(x_2 \lor \neg x_3)$.
In this work, we use SAT to denote both the Boolean Satisfiability problem and a satisfiable instance, which should be clear from the context.
We use unSAT to denote unsatisfiable instances.

There are many types of SAT solvers.
We focus on CDCL solvers, \minisat{}~\citep{sorensson2005minisat} in particular, because it is an open-source,  minimal, but powerful implementation.
A CDCL solver repeats the following steps: every iteration it chooses a variable and assigns it a binary value.
This is called a decision.
Then, the solver simplifies the formula building an implication graph and checks whether a conflict emerged.
Given a conflict, the solver can infer (learn) new clauses and backtrack to the variable assignments where the newly learned clause becomes unit (consisting of a single literal).
Learnt clauses force a variable assignment which avoids the previous conflict.
Sometimes, CDCL solver undoes all the variable assignments keeping the learned clauses to escape futile regions of the search space.
This is called a restart.

We focus on the branching heuristic because it is one of the most heavily used during the solution procedure.
The branching heuristic is responsible for picking the variable and assigning some value to it.
VSIDS~\citep{moskewicz2001chaff} is one of the most used CDCL branching heuristics.
It is a counter-based heuristic that keeps a scalar value for each literal or variable (\minisat{} uses the latter).
These values are increased every time a variable is involved in a conflict.
The algorithm behaves greedily with respect to these values called \emph{activities}.
Activities are usually initialized with zeroes~\citep{liang2015understanding}.

\subsection{Reinforcement Learning}
\label{sec:RL}

We formulate the RL problem as a Markov decision process (MDP).
An MDP is a tuple $\langle \mathcal{S}, \mathcal{A}, \mathcal{R}, \mathcal{T}, \gamma, \rho \rangle$ with a set of states $\mathcal{S}$, a set of actions $\mathcal{A}$, a reward function $\mathcal{R}(s,a,s')$  and the transition function $\mathcal{T}(s,a,s')=p(s,a,s')$, where $p(s,a,s')$ is a probability distribution, $s,s' \in \mathcal{S},\; a \in \mathcal{A}$. 
Discount factor $\gamma \in [0,1)$ weights preferences for immediate reward relative to future reward.
The last element of the tuple $\rho$ is the probability distribution over initial states.
In the case of \emph{episodic tasks}, the state space is split into the set of non-terminal states and the terminal state $\mathcal{S^{+}}$.
To solve an MDP means to find an optimal policy, a mapping that outputs an action or distribution over actions given a state and which maximizes the expected discounted return $R = \mathbb{E} [\sum_{t=0}^{\infty}\gamma^t r_t]$, where $r_t = \mathcal{R}(s_t, a_t, s_{t+1})$ is the reward for the transition from $s_t$ to $s_{t+1}$.
In Section~\ref{sec:method} we apply DQN, a value-based RL algorithm that approximates an optimal $Q$-function, an action-value function that estimates the sum of future rewards after taking an action $a$ in state $s$ and following an optimal policy $\pi$ thereafter: $Q^{*}(s,a) = \mathbb{E}_{\pi, \mathcal{T}, \rho}[\mathcal{R}(s,a,s') + \gamma \max_{a'} Q^*(s',a')]$. A mean squared temporal difference (TD) error is used to make an update step: $L(\theta) = (Q_{\theta}(s,a) - r - \gamma \max_{a'} Q_{\bar{\theta}}(s', a'))^2$, where $\theta$ parametrizes the $Q$-function. 
Target network~\citep{mnih2015human} $Q_{\bar{\theta}}$ is used to stabilize DQN.
Its weights are copied from the main network $Q_{\theta}$ after each $k$ minibatch updates.

\subsection{Graph Neural Networks}

Boolean formulas can be of arbitrary size.
Moreover, during the solution procedure, some parts of the formula are eliminated and new clauses are added.
We need a network architecture which does not assume the input to be of fixed size.
Moreover, a Boolean formula should be invariant to the permutation of the clauses, variables and their renaming. 
To accommodate these requirements and also to take the problem structure into account, we use Graph Neural Networks~\citep[GNN]{gori2005new} to approximate our $Q$-function.
We use the formalism of \citet{battaglia2018relational}, which unifies most existing GNN approaches.
Under this formalism, GNN is a set of functions that take an annotated graph as input and output a graph with modified annotations but the same topology.

Here, a graph is a directed graph $\langle V,E,U \rangle$, where $V$ is the set of vertices, $E$ is the set of directed edges with $e_{ij} = (i,j) \in E, \;v_i,v_j \in V$, and $U$ is a global attribute which contains the information relevant to the whole graph.
We call vertices, edges, and the global attribute entities.
Each entity has an associated annotation: $\vec e_{ij} \in \mathbb{R}^e$, $\vec v_i  \in \mathbb{R}^v$ or $\vec u \in \mathbb{R}^u$. 
A GNN changes these annotations as a result of its operations.

A GNN is as a set of six functions: update functions $\phi_e, \phi_v, \phi_u$ and aggregation functions $\rho_{e\rightarrow v}, \rho_{e \rightarrow u}, \rho_{v \rightarrow u}$.
The information propagates between vertices along graph edges.
Update functions compute new entity annotations.
Aggregation functions enable GNN to process graphs of arbitrary topology, compressing multiple entities features into vectors of fixed size.
Summation, averaging, taking $\max$ or $\min$ are popular choices of aggregation functions.

More formally, within one iteration, a GNN does the following computations (in order):

\begin{equation*}
\centering
\begin{array}{ll}
     \vec e'_{ij} &= \phi_e(\vec u, \vec e, \vec v_i, \vec v_j) \;\forall e_{ij} \in E  \\[3pt]
     \vec v'_{i} &= \phi_v\big[\vec u, \vec v_i, \rho_{e\rightarrow v}(\{\vec e_{ki} \;|\; \forall e_{ki} \in E\})\big] \;\forall v_i \in V \\[3pt]
     \vec u' &= \phi_u\big[\vec u, \rho_{e \rightarrow u}(\{\vec e_{ij} \;|\; \forall e_{ij} \in E\}), \rho_{v \rightarrow u}(\{\vec v_{i} \;|\; \forall v_{i} \in V\})\big].
\end{array}
\end{equation*}

A GNN performs multiple iterations to further propagate information in the graph. 
Neural networks that represent update functions, can be optimised end-to-end using backpropagation.

%% file: sections/method.tex
\section{\method{}}
\label{sec:method}

As noted in Section~\ref{sec:RL}, we use the MDP formalism for our purposes.
Each SAT problem is an MDP sampled from a distribution of SAT problems of a specific family (e.g., random 3-SAT or graph coloring). 
Moreover, each problem is either satisfiable or unsatisfiable.
Hence, a task is defined as follows: $\tau \sim \mathcal{D}(\phi, (un)SAT, n_{vars}, n_{clauses})$, where $\mathcal{D}$ is the distribution of SAT problems with $\phi$ defining the task family, the second argument defining the problem satisfiability and $n_{vars}$ and $n_{clauses}$ are the number of variables and clauses respectively.
Each state of the MDP consists of unassigned variables and unsatisfied clauses containing these variables.
The MDP is episodic, and a terminal state is reached when a satisfying assignment is found, or the all possible options have been exhausted, proving unSAT.
The action space includes two actions for each unassigned variable: assigning it to true or false.
We build upon the \minisat{}-based environment of \citep{wang2018gameplay}. 
We modified it to support arbitrary SAT problems and generate a graph representation of the state.
It takes the actions, modifies its implication graph internally and returns a new state containing newly learned clauses and without the variables removed during propagation.
Strictly speaking, this state is not fully observable. 
In the case of a conflict, the solver undoes the assignments for variables that are not observed by the agent.
However, in practice, this should not inhibit the goal of quickly pruning the search tree:
the information in the state is enough to pick a variable that leads to more propagation in the remaining formula.
We use a simple reward function: the agent gets a negative reward of $p$ for each non-terminal transition and $0$ for reaching the terminal state.
This reward encourages an agent to finish an episode as quickly as possible and does not require elaborate reward shaping.

SAT is an appealing problem from the RL perspective. 
It has the features that are hard to find in conventional RL environments.
First, elements of the state/action set are of different dimensions, which is a challenging case for conventional function approximation techniques.
Second, the state-space has a structured, object-oriented representation.
Third, SAT allows to vary problem sizes without changing the task family, and to change the task family without changing problem sizes.
Lastly, with~\citet{hoos2000satlib} benchmarks we use, experiments do not take weeks and are easy to iterate on.

\subsection{State Representation}
\begin{figure}
\centering
\begin{minipage}{.49\textwidth}
  \centering
\begin{tikzpicture}
        \Vertex[label=$x_1$,size=0.8, color=orange, y=2]{x1}
        \Vertex[label=$x_2$,size=0.8, color=orange, y=1]{x2}
        \Vertex[label=$x_3$,size=0.8, color=orange]{x3}
        \Vertex[label=$c_1$,size=0.8, color=orange, x=2.2, y=1.5]{c1}
        \Vertex[label=$c_2$,size=0.8, color=orange, x=2.2, y=0.5]{c2}
        \Text[x=-0.8, y=2.2, anchor = north east]{[1,0]}
        \Text[x=-0.8, y=1.2, anchor = north east]{[1,0]}
        \Text[x=-0.8, y=0.2, anchor = north east]{[1,0]}
        
        \Text[x=2.8, y=1.5, anchor = west]{[0,1]}
        \Text[x=2.8, y=0.5, anchor = west]{[0,1]}
        
        \Edge[Direct, label=$0\text{,}1$](x1)(c1)
        \Edge[Direct, label=$0\text{,}1$](c1)(x1)
        \Edge[Direct,label=$0\text{,}1$](x2)(c1)
        \Edge[Direct,label=$0\text{,}1$](c1)(x2)
        \Edge[Direct,label=$1\text{,}0$](x2)(c2)
        \Edge[Direct,label=$1\text{,}0$](c2)(x2)
        \Edge[Direct,label=$0\text{,}1$](x3)(c2)
        \Edge[Direct,label=$0\text{,}1$](c2)(x3)
    \end{tikzpicture}
\caption{Bipartite graph representation of the Boolean formula $(x_1 \lor x_2) \land (\neg x_2 \lor x_3)$. The numbers next to the vertices distinguish variables and clauses. Edge labels encode literal polarities.}
\label{fig:state}
\end{minipage}%
\hfill
\begin{minipage}{.49\textwidth}
  \centering
\begin{tikzpicture}
        \Vertex[label=$x_1$,size=0.8, color=orange, y=2]{x1}
        \Vertex[label=$x_2$,size=0.8, color=orange, y=1]{x2}
        \Vertex[label=$x_3$,size=0.8, color=orange]{x3}
        \Vertex[label=$c_1$,size=0.8, color=orange, x=2.2, y=1.5]{c1}
        \Vertex[label=$c_2$,size=0.8, color=orange, x=2.2, y=0.5]{c2}
        \Text[x=-0.8, y=2.2, anchor = north east]{[42.0 , 3.14]}
        \Text[x=-0.8, y=1.2, anchor = north east]{[1.62 , 2.70]}
        \Text[x=-0.8, y=0.2, anchor = north east]{[6.02 , 1.67]}
        \Edge[Direct](x1)(c1)
        \Edge[Direct](x2)(c1)
        \Edge[Direct](x2)(c2)
        \Edge[Direct](x3)(c2)
        \Edge[Direct](c1)(x1)
        \Edge[Direct](c1)(x2)
        \Edge[Direct](c2)(x2)
        \Edge[Direct](c2)(x3)
    \end{tikzpicture}
\caption{$Q$-function values for setting variables to \emph{true} and \emph{false} respectively. Taking $\arg\max$ across all $Q$-values of variable nodes gives an action.}
% Since GNN work on directed graphs, we add two directed edges to connect two nodes.
\label{fig:act}
\end{minipage}
\end{figure}
We represent a SAT problem as a graph similar to~\citet{selsam2018learning}.
We make it more compact, using vertices to denote variables instead of literals.
We use vertices to encode clauses as well.
As Figure~\ref{fig:state} shows, our state representation is simple and does not require feature engineering.
An edge $(x_i, c_i)$ means that a clause $c_i$ contains literal $x_i$.
If a literal contains a negation, a corresponding edge has a $[1,0]$ label and $[0,1]$ otherwise.
GNNs process directed graphs, so we create two directed edges with the same labels: from a variable to a clause and vice-versa.
Vertex features are two-dimensional one-hot vectors, denoting either a variable or a clause.
We do not provide any other information to the model. 
The global attribute input is empty and is only used for message passing.
\subsection{$Q$-Function Representation}
We use the encode-process-decode architecture \citep{battaglia2018relational}, which we discuss in more detail in Appendix~\ref{sec:arch}.
Similarly to~\citet{bapst2019structured}, our GNN labels variable vertices with $Q$-values. 
Each variable vertex has two actions: set the variable to true or false as shown on Figure~\ref{fig:act}.
We choose the action that gives the maximal $Q$-value across all variable vertices.
The graph contains only unassigned variables, so all actions are valid.
We use DQN with common techniques such as memory replay, target network, and $\epsilon$-greedy exploration. 
To expose the agent to more episodes and prevent it from getting stuck, we cap the maximum number of actions per episode similarly to the \emph{episode length} parameter in \emph{gym}~\citep{openaigym}.
We implement our models using Pytorch~\citep{paszke2017automatic} and Pytorch Geometric~\citep{Fey/Lenssen/2019}.
\subsection{Training and Evaluation}
We train our agent using Random 3-SAT instances from the SATLIB benchmark~\citep{hoos2000satlib}.
To measure generalization, we split these data into training, validation, and test sets.
To illustrate the problem complexities, Table~\ref{tab:data} provides the number of steps it takes \minisat{} to solve the problem.
Each random 3-SAT problem is denoted as SAT-X-Y or unSAT-X-Y, where SAT means that all problems are satisfiable, unSAT means all problems are unsatisfiable. 
X and Y stand for the number of variables and clauses in the initial formula.
We provide more details about the datasets in Appendix~\ref{sec:dataset}.

While random 3-SAT problems have relatively few variables and clauses, they have an interesting property that makes them more challenging for a solver.
For this dataset, the ratio of clauses to variables is close to 4.3 to 1 which is near the \emph{phase transition} at which it is hard to say whether the problem is SAT or unSAT~\citep{cheeseman1991really}.
In 3-SAT problems, each clause has exactly 3 variables. However, learned clauses might be of arbitrary size.

We use \fullmetric{} w.r.t.\ \minisat{} as our main performance metric: the number of iterations it takes \minisat{} to solve a problem divided by \method{}'s number of iterations.
Similarly to the \emph{median human normalized score} adopted in the Atari domain~\citep{hessel2018rainbow}, we use the median instead of the mean to avoid skew from outliers.
By one iteration we mean one \emph{decision}, i.e., choosing a variable and setting it to a value.
We compare ourselves with the best \minisat{} results having run \minisat{} with and without restarts.
We cap the number of decisions our method takes at the beginning of the solution procedure and then we give control to \minisat{}.

We are not interested in the absolute number of iterations per se or the total ratio between VSIDS and Graph-Q-SAT. We use these numbers as a common scale to show the generalisation, transfer and data efficiency properties of our approach.

When training, we evaluate the model every 1000 batch updates on the validation instances and pick the model with the best validation results.
After that, we evaluate this model on the test set and report the results.
For each model we do 5 training runs and report the average \metric{} results, the maximum, and the minimum. 
\begin{table}[tb]
    \begin{minipage}{.45\linewidth}
    \caption{Number of \minisat{} iterations (no restarts) to solve random 3-SAT instances.}
\centering
\begin{tabular}[t]{lll}
dataset & median & mean \\ \midrule
SAT 50-218 & 38 & 42 \\
SAT 100-430 & 232 & 286 \\
SAT 250-1065 & 62 192 & 76 120 \\
\midrule
unSAT 50-218 & 68 & 68 \\ 
unSAT 100-430 & 587 & 596 \\ 
unSAT 250-1065 & 178 956 & 182 799
\label{tab:data}
\end{tabular}
\end{minipage}%
\hfill
\begin{minipage}{.525\linewidth}
\centering
\caption{\method{} \metric{} trained on SAT-50-218. SAT-50-218 results are for a separate validation set.} 
\begin{tabular}[t]{lccc}
dataset & mean & min & max \\ \midrule
SAT 50-218 &  2.46&2.26&2.72   \\
SAT 100-430 & 3.94&3.53&4.41  \\
SAT 250-1065 & 3.91&2.88&5.22  \\
\midrule
unSAT 50-218 &  2.34&2.07&2.51    \\ 
unSAT 100-430 &2.24&1.85&2.66  \\
unSAT 250-1065 &1.54&1.30&1.64 
\label{tab:3sat}
\end{tabular}
    \end{minipage} 
\end{table}
We provide all the hyperparameters needed to reproduce our results in Appendix~\ref{sec:reproducibility}.
Our experimental code as well as the \minisat{} \emph{gym} environment can be found at \url{https://github.com/NVIDIA/GraphQSat}.

%% file: sections/experiments.tex
\section{Experimental Results}

In this section, we present empirical results for \method{}.

\subsection{Improving upon VSIDS}
\label{sec:vsids}
In our first experiment, we consider whether it is possible to improve upon VSIDS using no domain knowledge, a simple state representation, and a simple reward function.
The first row in Table~\ref{tab:3sat} gives a positive answer to that question.
DQN equipped with a GNN solves the problems in fewer than half the iterations of \minisat{}.
\method{} makes decisions resulting in more propagations, i.e., inferring variable values based on other variable assignments and clauses.
This helps \method{} prune the search tree faster.
For SAT-50-218, \method{} does on average 2.44 more propagations than \minisat{} (6.62 versus 4.18).
We plot the average number of variable assignments for each problem individually in the Appendix~\ref{sec:propagations}.

These results raise the question: Why does \method{} outperform VSIDS? 
VSIDS is a counter-based heuristic that takes time to warm up.
Our model, on the other hand, perceives the whole problem structure and can make more informed decisions from the beginning.
To test this hypothesis, we vary the number of decisions our model makes at the beginning of the solution procedure before we hand the control back to VSIDS.
The results in Figure~\ref{fig:k_steps} support this hypothesis.
Even if our model is used for only the first ten iterations, it still improves performance over VSIDS.

One strength of \method{} is that VSIDS keeps being updated while the decisions are made with \method{}. 
We believe that \method{} complements VSIDS by providing better quality decisions in the initial phase while VSIDS is warming up.
Capping the number of model calls also significantly reduces the main bottleneck of our approach -- wall clock time spent on model evaluation.

\subsection{Generalization Properties of \method{}}

Next, we consider \method{}'s generalization properties.

\subsubsection{Generalization across Problem Sizes}
\label{sec:sizegen}

Table~\ref{tab:3sat} shows that \method{} has no difficulty generalizing to larger problems, showing almost 4X improvement in iterations for a dataset 5 times bigger than the training set.
\method{} on average leads to more variable assignments changes per step, e.g., 7.58 vs\ 5.89 on SAT-100-430 (refer to Appendix~\ref{sec:propagations} for detailed plots).
It might seem surprising that the model performs better for larger problems.
However, an increase in score for different problem sizes might also mean that the base solver scales worse than our method does for this benchmark.

\subsubsection{Generalization from SAT to unSAT}
\label{sec:sat2unsat}

An important characteristic of \method{} is that the problem formulation and representation makes it possible to solve unSAT problems when training only on SAT, which is problematic for some existing approaches~\citep{selsam2018learning}.
The performance is, however, worse than the performance on satisfiable problems.
On the one hand, SAT and unSAT problems are different.
When the solver finds one satisfying assignment, the problem is solved.
For unSAT, the algorithm needs to exhaust all possible options to prove that there is no such assignment.
On the other hand, there is one important similarity between the two: an algorithm has to prune the search tree as fast as possible.
Our measurements of the average number of propagations per step demonstrate that \method{} learns how to prune the tree more efficiently than VSIDS (6.36 vs 4.17 for unSAT-50-218, detailed plots are in Appendix~\ref{sec:propagations}).

\subsubsection{Transfer across Task Families}

So far, we have examined the generalization properties of \method{} varying only the last three arguments of the task distribution defined in Section~\ref{sec:method} ($\mathcal{D}(\phi, (un)SAT, n_{vars}, n_{clauses})$.
In this section we go one step further and study \method{}'s \textit{zero-shot transfer} to a new task family $\phi$. 

This is a challenging problem.
SAT problems have distinct structures, e.g.,
the graph representation of a random 3-SAT problem looks different than that of a graph coloring problem.
GNNs learn graph local properties, i.e. how neighbouring entities' features have a global implication on $Q$-values. 
It is reasonable to expect a performance drop when changing the task family $\phi$, but the magnitude of the drop gives some indication of the method's ability to transfer across task families.
Therefore, we evaluate a model trained on SAT-50-218 on the flat graph coloring benchmark from SATLIB~\citep{hoos2000satlib}.
All the problems in the benchmark are satisfiable.
\begin{figure}
\centering
\begin{minipage}{.49\textwidth}
\centering
        \includegraphics[height=120pt]{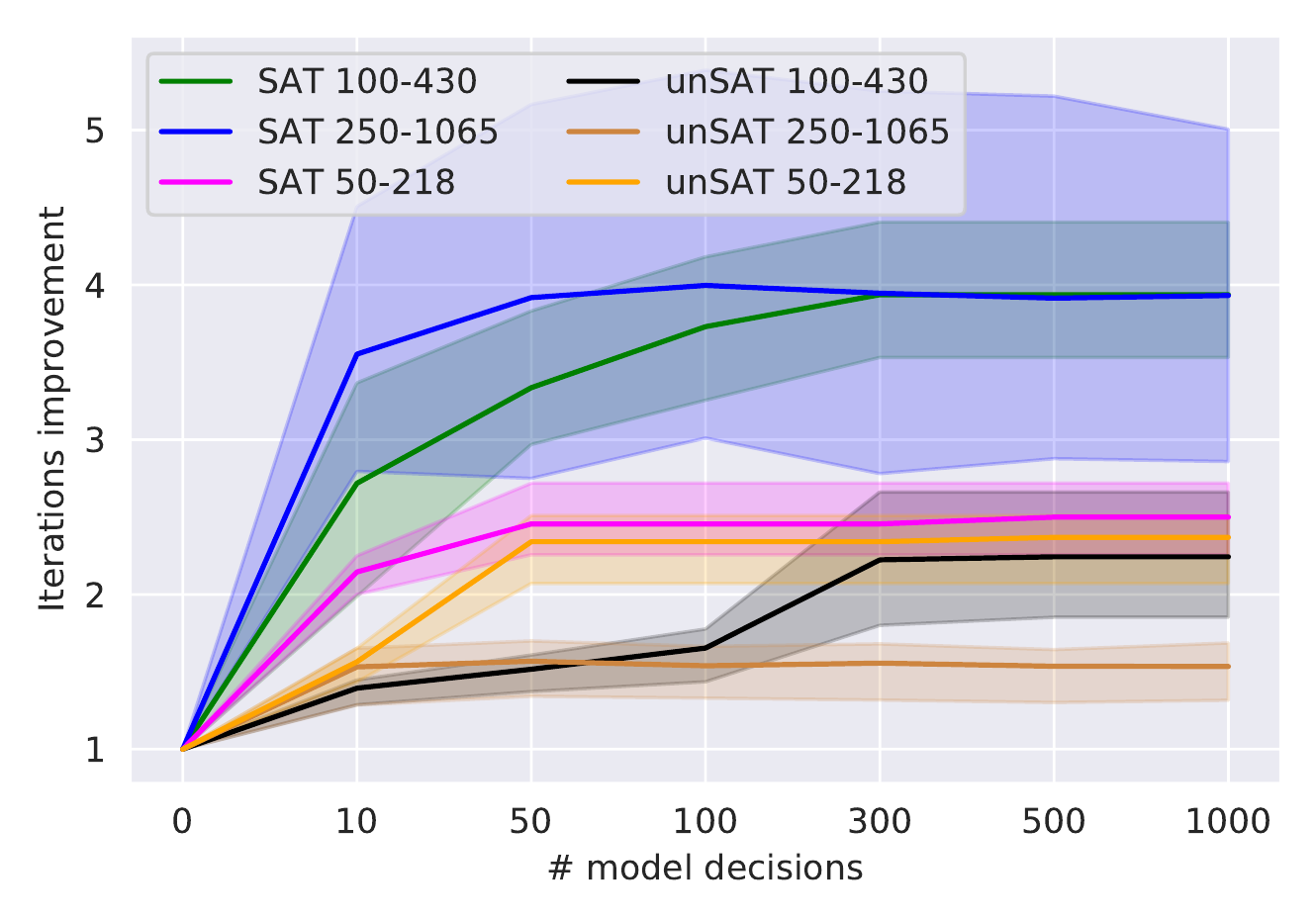}
        \caption{\method{} number of maximum first decisions vs performance. \method{} shows improvement starting from 10 iterations confirming our hypothesis of VSIDS initialization problem. The shades mark \textit{min} and \textit{max} values.}
        \label{fig:k_steps}
\end{minipage}%
\hfill
\begin{minipage}{.49\textwidth}
\centering
\includegraphics[height=120pt]{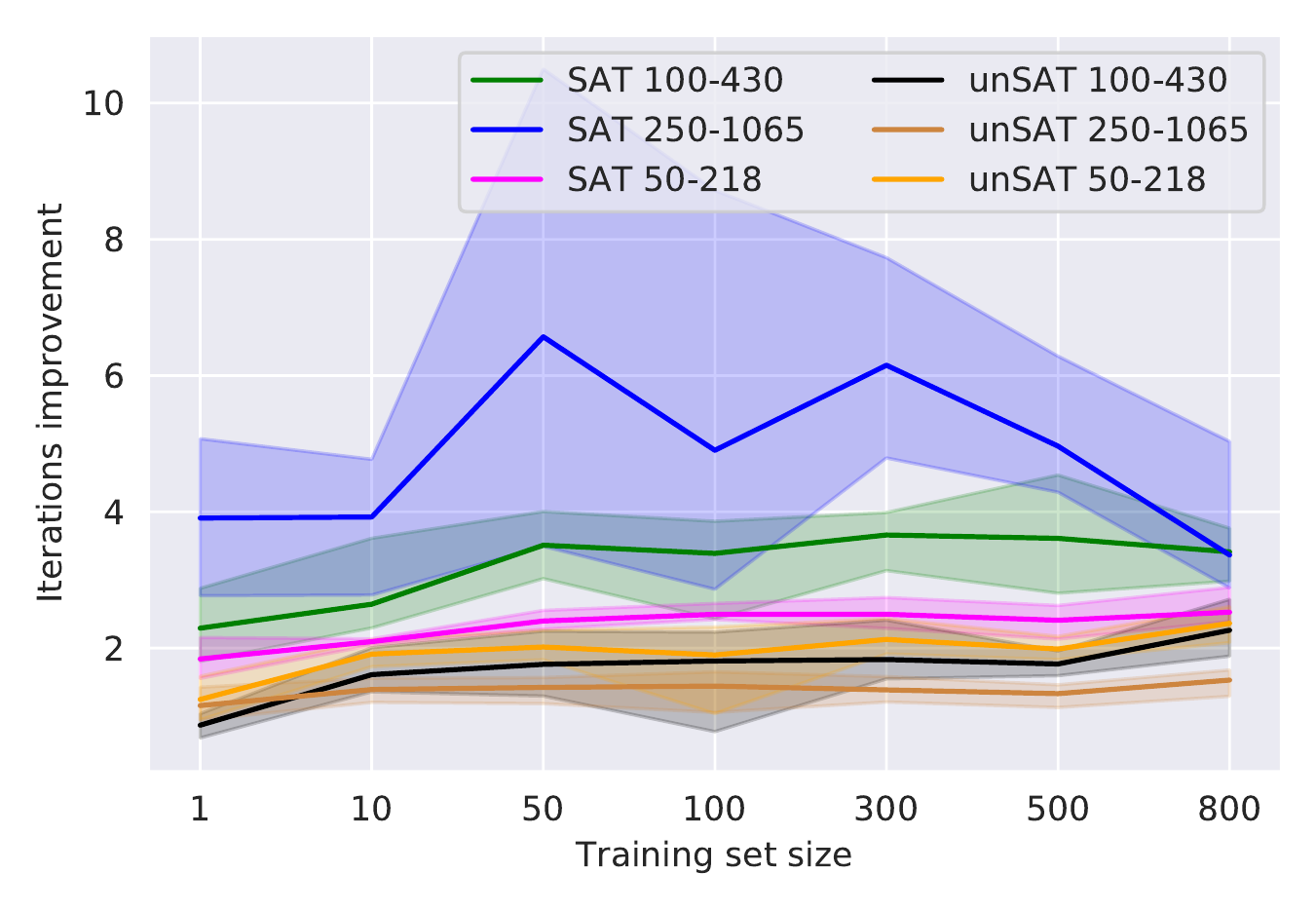}
\caption{Dataset size effect on generalization. While \method{} profits from more data in most of the cases, it is able to generalize even from one data point. Model is trained on SAT-50-218. The shades mark \textit{min} and \textit{max} values.}
\label{fig:data_efficiency}
\end{minipage}
\end{figure}

Table~\ref{tab:gen2color} shows positive transfer for \method{} on the graph coloring benchmark, with MRIR above 1 in five out of eight cases.
As expected, \metric{} is lower if than in Table~\ref{tab:3sat}, where the model was evaluated on the tasks sampled from the same distribution.

Training directly on the graph coloring benchmark indeed improves performance.
Graph coloring benchmarks have only 100 problems each, so we do not split them into training/validation/test sets using \emph{flat-75-180} for training and \emph{flat-100-239} to do model selection.
Table~\ref{tab:structured} shows that \method{}, trained on flat75-180 shows higher \metric{} compared to the transferred model. 
Additionally, this experiment shows that \method{} can scale when training on larger graphs.

Another intriguing property of \method{} generalization is that sometimes \method{} shows better performance when generalizing in comparison to training from scratch.
Learning on SAT-100-430 requires more resources, does not generalize as well, and is generally less stable than training on SAT-50-218 and then transferring to SAT-100-430 and SAT-250-1065.
Hence, generalizing with \method{} not only reduces the time and samples spent on training, but yields models hardly achievable by learning.
We suppose the reason is that transfer is not directly affected by all the issues an RL agent faces when training: higher variance in the returns caused by longer episodes, challenges for temporal credit assignment, and difficulties with exploration.

\subsection{Data Efficiency}

We design our next experiment to understand how many different SAT problems \method{} needs to learn from.  
We varied the SAT-50-218 training set from a single problem to 800 problems.
Figure~\ref{fig:data_efficiency} shows that \method{} is extremely data efficient.
Having more data helps in most cases but, even with a single problem, \method{} generalizes across problem sizes and to unSAT instances.
This should allow \method{} to generalize to new benchmarks without access to many problems from them.
We assume that \method{}'s data efficiency is one of the benefits of using RL.
The environment allows the agent to explore diverse regions of state-action space, making it possible to learn useful policies even from a single instance. 
In supervised learning, data diversity is addressed at the training data generation step. 

\subsection{Wall-Clock Time Bottleneck}
\label{sec:timings}
The main goal of this work is to show that RL can learn a value function that can be used as a branching heuristic in a SAT solver, and to study the model's generalisation properties.
In its current form, more work would be required to apply \method{} in an industrial setting, where wall-clock time is the metric of success, and problem sizes are extremely large.
However, we believe that \method{} should be of interest to the SAT community because reduction of iterations can reduce wall clock time when the number of saved iterations is large enough to tolerate the network inference timings.
Due to a shortage of space, we present the wall-clock time and scaling analysis in Appendix~\ref{sec:walltime}.
This analysis shows that MRIR reduction leads to wall clock time improvements on SAT-250 and unSAT-250. 
\begin{table}[t]
\centering
\centering
\begin{minipage}{.64\textwidth}
\caption{SAT-50 model's performance on SATLIB flat graph coloring benchmark. The comparison is w.r.t.\ \minisat{} with restarts, since \minisat{} performs better in this mode for this benchmark.}
\begin{tabular}{lcccccc}
\multirow{2}{*}{dataset} & \multirow{2}{*}{variables} & \multirow{2}{*}{clauses} & \multicolumn{3}{c}{\method{} \metric{}} \\
 &  &  &  \multicolumn{1}{c}{average} & \multicolumn{1}{c}{min} & \multicolumn{1}{c}{max} \\
 \midrule
30-60 &90&300& 1.51& 1.25&1.65\\
50-115 &150 &545& 1.36& 0.47&1.80\\
75-180 &225 &840& 1.40& 0.31&2.06\\
100-239 &300 &1117& 1.44&0.31&2.38\\
125-301 &375 &1403  & 1.02&0.32&1.87\\
150-360 &450&1680  & 0.76&0.37&1.40\\
175-417 &525  &1951 & 0.67&0.44&1.36\\
200-479 & 600 & 2237& 0.67&0.54&0.87
\label{tab:gen2color}
\vspace{5pt}
\end{tabular}
\end{minipage}
\hfill
\begin{minipage}{.34\textwidth}
\centering
\caption{\method{} \metric{} (5 training runs on 75-180, model selection with 100-239).}
\label{tab:structured}
\begin{tabular}{lccc}
\multirow{2}{*}{dataset} & \multicolumn{3}{c}{\method{} \metric{}} \\
 & \multicolumn{1}{c}{average} & \multicolumn{1}{c}{min} & \multicolumn{1}{c}{max} \\
\midrule
75-180 & 2.44 & 2.25 & 2.70 \\
100-239 & 2.89 & 2.77 & 2.98 \\
\midrule
30-60 & 1.74 & 1.33 & 2.00 \\
50-115 & 2.08 & 2.00 & 2.13 \\
125-301 & 2.43 & 2.20 & 2.66 \\
150-360 & 2.07 & 2.00 & 2.11 \\
175-417 & 1.98 & 1.69 & 2.21 \\
200-479 & 1.70 & 1.38 & 1.98
\end{tabular}
\end{minipage}
\end{table}

%% file: sections/related_work.tex
\section{Related Work}

Using machine learning for the SAT problem is not a new idea~\citep{grozea2014can, haim2009restart, flint2012perceptron, singh2009avatarsat, xu2008satzilla, liang2016learning}.
Recently, SAT has attracted interest in the deep learning community. There are two main approaches: solving a problem end-to-end or learning heuristics while keeping the algorithm backbone the same.
\citet[NeuroSAT]{selsam2018learning} take an end-to-end supervised learning approach demonstrating that GNN can generalize to SAT problems bigger than those used for training.
NeuroSAT finds satisfying assignments for the SAT formulae and thus cannot generalize from SAT to unSAT problems.
Moreover, the method is incomplete and might generate incorrect results, which is extremely important, especially for unSAT problems.
\citet{selsam2019guiding} modify NeuroSAT and integrate it into popular SAT solvers to improve timing on SATCOMP-2018 benchmark.
While the approach shows its potential to scale to large problems, it requires an extensive training set including over 150,000 data points.
\citet{amizadeh2018learning} propose an end-to-end GNN architecture to solve circuit-SAT problems. While their model never produces false positives, it cannot solve unSAT problems.

The following methods take the second approach: learning a branching heuristic instead of learning an algorithm end-to-end.
\citet{jaszczur2019heuristics} take the supervised learning approach using the same graph representation as~\citet{selsam2018learning}.
The authors show a positive effect of combining DPLL/CDCL solver with the learnt model.
As in~\citet{selsam2018learning}, their approach requires diligent crafting of the test set.
Also, the authors do not compare their approach to the VSIDS heuristic, which is known to be a crucial component of CDCL~\citep{katebi2011empirical}.
\citet{wang2018gameplay}, whose environment we took as a starting point, show that DQN does not generalize for 20-91 3-SAT problems, whereas Alpha(Go) Zero~\citep{SilverSSAHGHBLB17} does.
Our results show that the issue is related to state representation.
They use CNNs, which are not invariant to variable renaming or permutations.
Moreover, CNNs require a fixed input size which makes it infeasible when applying to problems with different numbers of variables or clauses.

\citet{yolcu2019learning} use REINFORCE~\citep{williams1992simple} to learn the variable selection heuristic of a local search SAT solver~\citep{selman1993local}. 
Their algorithm is an incomplete solver and cannot work with unsatisfiable instances.
They also investigate the generalisation over problem sizes on random instances near the phase transition.
However, in this experiment, the training problems have ten variables only, and the number of variables in the test set does not exceed 80 with the success ratio of the algorithm staying below the baseline for the latter case.

\citet{lederman2020learning} train a REINFORCE~\citep{williams1992simple} agent applying GNNs to replace the branching heuristic for Quantified Boolean Formulas (QBF).
QBF considers a different problem allowing existential and universal quantifiers.
\citet{lederman2020learning} note positive generalization properties across problem size for problems from similar distributions.
Our work focuses more on the generalization and transfer properties of a GNN value-based RL algorithm.
We investigate data efficiency properties and merge VSIDS with a trained RL agent, looking into the trade-off between the model use and its effect on the final solution.
Apart from that, we show that it is possible to achieve good performance and generalization properties with a simpler state representation.
Finally, doing more message propagations per step and using a GNN as a $Q$-function (in their case, a GNN only computes node embeddings) allows us to consider more subtle dependencies in the graph.

Look-ahead SAT solvers~\citep{heule2009look} perform more computations compared to VSIDS to evaluate the consequences of their decisions. 
In some of the cases, e.g. random $k$-SAT, this pays off.
Difference heuristics used for making a decision measure reduction in the formulae before and after the decision.
LRB heuristic~\citep{liang2016learning} uses multi-armed bandits to explicitly optimise for the ability of the variables' to generate learnt clauses.
We hypothesise, that~\method{} might have learnt some aspects of those heuristics (Figure~\ref{fig:props-per-step} in Appendix~\ref{sec:propagations}).
We believe that integrating \method{} with other types of solvers is a promising direction for future research.

\citet{vinyals2015pointer} introduce a recurrent architecture for approximately solving complex problems, such as the Traveling Salesman Problem, approaching it in a supervised way.
\citet{bello2016neural} consider combinatorial optimization problems with RL. 
\citet{khalil2017learning} approach combinatorial optimization using GNNs and DQN, learning a heuristic that is later used greedily.
It differs from our approach in that their heuristic is effectively the algorithm itself.
The environment dynamics in~\citet{khalil2017learning} is straightforward with the next state easily inferred, given the current state and the chosen action. In the case of SAT, there are CDCL steps after the decision, and the next state might be totally different from the current one making the problem harder in terms of learning the $Q$-function.
In addition, we use~\citet{battaglia2018relational} which is more expressive than \texttt{structure2vec} used in~\citet{khalil2017learning}. 
The global attribute in~\citet{battaglia2018relational} can facilitate message passing in case of a bigger graph. 
Having separate updaters for edges and nodes leads to more powerful representations. 
And, finally, an edge updater of~\citet{battaglia2018relational} can learn better pairwise interaction between the sender and the receiver, enabling sending different messages to different nodes.

\citet{paliwal2019graph} use GNNs with imitation learning for theorem proving.
\citet{carbune2018smartchoices} propose a general framework of injecting an RL agent into existing algorithms. 
\citet{cai2019reinforcement} use RL to find a suboptimal solution that is further refined by another optimization algorithm, in their case, simulated annealing~\citep[SA]{kirkpatrick1983optimization}.
It is not restricted to SA, and this modularity is valuable.
However, it is also a drawback because the second optimization algorithm might benefit more from the first if they were interleaved.
For instance, \method{} can guide search before VSIDS overcomes its initialization bias.

GNNs have enabled the study of RL agents in state/action spaces of dynamic size, which is crucial for generalization beyond the given task.
\citet{wang2018nervenet} and \citet{sanchez2018graph} consider GNNs for the control problem generalization.
\citet{bapst2019structured} report strong generalization capabilities for the construction task.
Multi-agent research~\citep{jiang2018graph, aleks2018deep, agarwal2019learning} shows that GNN benefits from invariance to the number of agents in the team or other environmental entities.

%% file: sections/discussion.tex
\section{Conclusions and Future Work}

In this paper, we demonstrated that $Q$-learning can be used to learn the branching heuristic of a SAT solver.
\method{} uses a simple state representation and does not require elaborate reward shaping.
We show empirically that \method{} causes more variable propagations per step, solving the SAT problem in fewer iterations than VSIDS.
For larger problems, we showed that fewer iterations could, in turn, reduce wall-clock time.%, a key indicator of \method{}'s potential.
We demonstrated its generalization abilities, showing more than 2-3X reduction in iterations for problems up to 5X larger and 1.5-2X from SAT to unSAT.
We showed how \method{} improves VSIDS and that it is data-efficient.
We also demonstrated positive transfer properties when changing the task family and showed that training on data from other distributions could lead to further performance improvements.

Although we showed the powerful generalization properties of graph-based RL on SAT, we believe the problem is still far from solved. More work is needed before \method{} is ready to compete with branching heuristics in a modern industrial setting. 
The two main direction of future applied research are scaling and wall-clock time reduction.
Some possible ways of tackling these issues include combining the machine learning improvements from above together with an efficient C++ implementation, using a smaller network, reducing the network polling frequency, and replacing the variable activities with \method{}'s output, similarly to~\citet{selsam2019guiding}.

From the machine learning perspective, it is intriguing to study how combining benchmarks from different domains might improve the transfer behavior.
Further research will focus on scaling \method{} using the latest stabilizing techniques~\citep{hessel2018rainbow} and more sophisticated exploration methods.
Building an efficient curriculum is another important step towards further scaling the method, motivated by \citet{bapst2019structured}. % who show a positive effect of curriculum learning on RL with GNN.
\citet{newsham2014impact} show that the graph structure of SAT problems affects the problem complexity.
We are interested in understanding how the structure influences the performance of~\method{} and how we can exploit this knowledge to improve \method{}.

%% file: sections/impact.tex
\section*{Broader Impact}

We believe that further progress in machine learning can have a profound economic, societal and political impact.
It is hard to predict a particular effect of our method on society but, in general, we believe that the society might benefit from our research through its impact on industry and academia.
We consider two examples below.

SAT has a profound impact on circuit design, computer security, artificial intelligence, automatic theorem proving, and combinatorial optimisation, among others.
For academia, \method{} code and results give a playground to work on GNN scaling, generalisation in RL, transfer and multitask learning, and incorporating a machine learning model with a well established algorithm. 
It can encourage collaboration between the applied ML and SAT communities.
Analysing the behaviour of learned models might give human designers more insights to boost further research.

For industry, having faster SAT solvers would lead to faster production cycles and faster rate of progress as well as to more robust products.
In circuitry design, for example, SAT is used for hardware verification.
As a result, faster SAT solvers will eventually lead to fewer faults in hardware.

Like any technology, our method also carries potential risks.
Further automation might reduce the need for human labour.
If not managed and regulated properly, machine learning progress might also exacerbate social and economic inequality.

%% file: sections/acknowledgements.tex
\begin{ack}
The authors would like to thank Rajarshi Roy, Robert Kirby, Yogesh Mahajan, Alex Aiken, Mohammad Shoeybi, Rafael Valle, Sungwon Kim and the rest of the Applied Deep Learning Research team at NVIDIA for useful discussions and feedback.
The authors would also like to thank Andrew Tao and Guy Peled for providing computing support.
The authors thank Henry Kenlay for useful discussions on GNN scaling properties.

Vitaly Kurin is a doctoral student at the University of Oxford funded by Samsung R\&D Institute UK through the \emph{Autonomous Intelligent Machines and Systems} program. 
This work was done during Vitaly's internship at NVIDIA.
Saad Godil is a Director of Applied Deep Learning Research at NVIDIA. 
Shimon Whiteson is a Professor of Computer science at the University of Oxford and the Head of Research at Waymo UK. 
Shimon Whiteson has received funding from the European Research Council under the European Union’s Horizon 2020 research and innovation programme (grant agreement number 637713).
Bryan Catanzaro is a Vice President, Applied Deep Learning Research at NVIDIA. 
\end{ack}

%% file: sections/appendix.tex
\clearpage

\section{Wall-clock Time and Scaling Analysis}
\label{sec:walltime}

\begin{figure}[htp]
  \centering
  \subfigure[SAT-250]{\includegraphics[width=0.45\columnwidth]{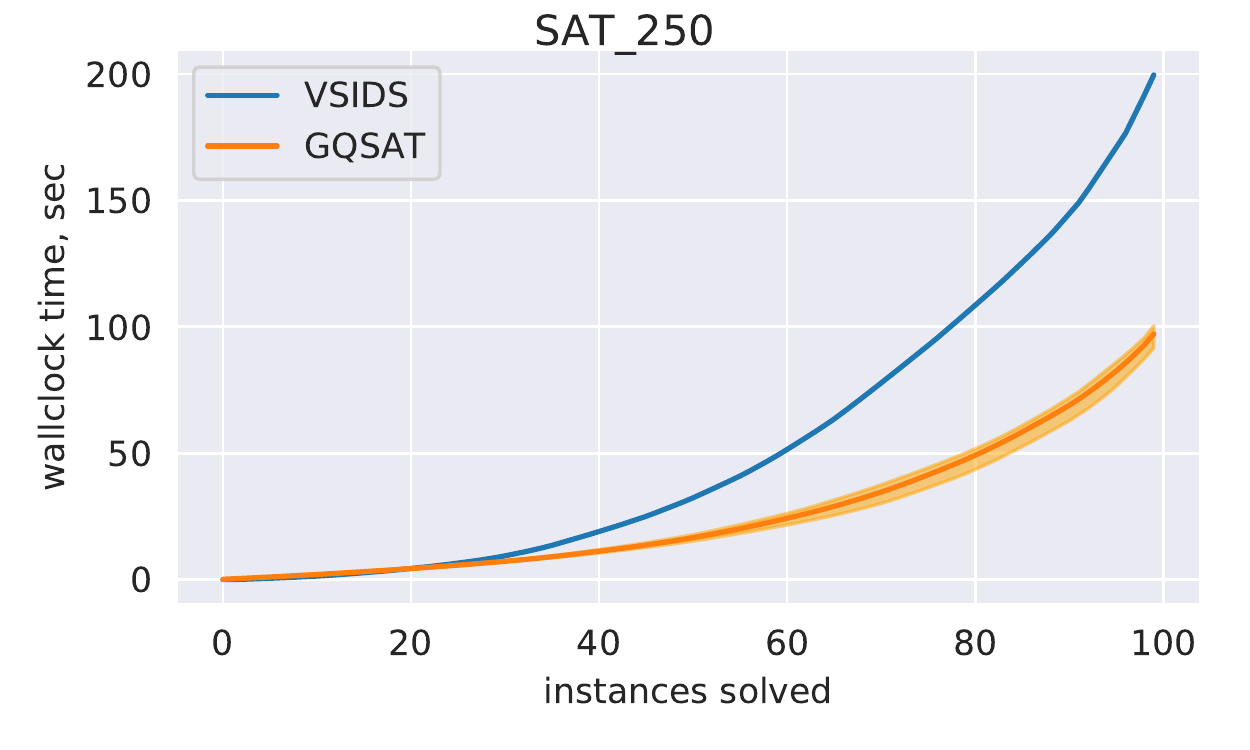}}\hfill
  \subfigure[unSAT-250]{\includegraphics[width=0.45\columnwidth]{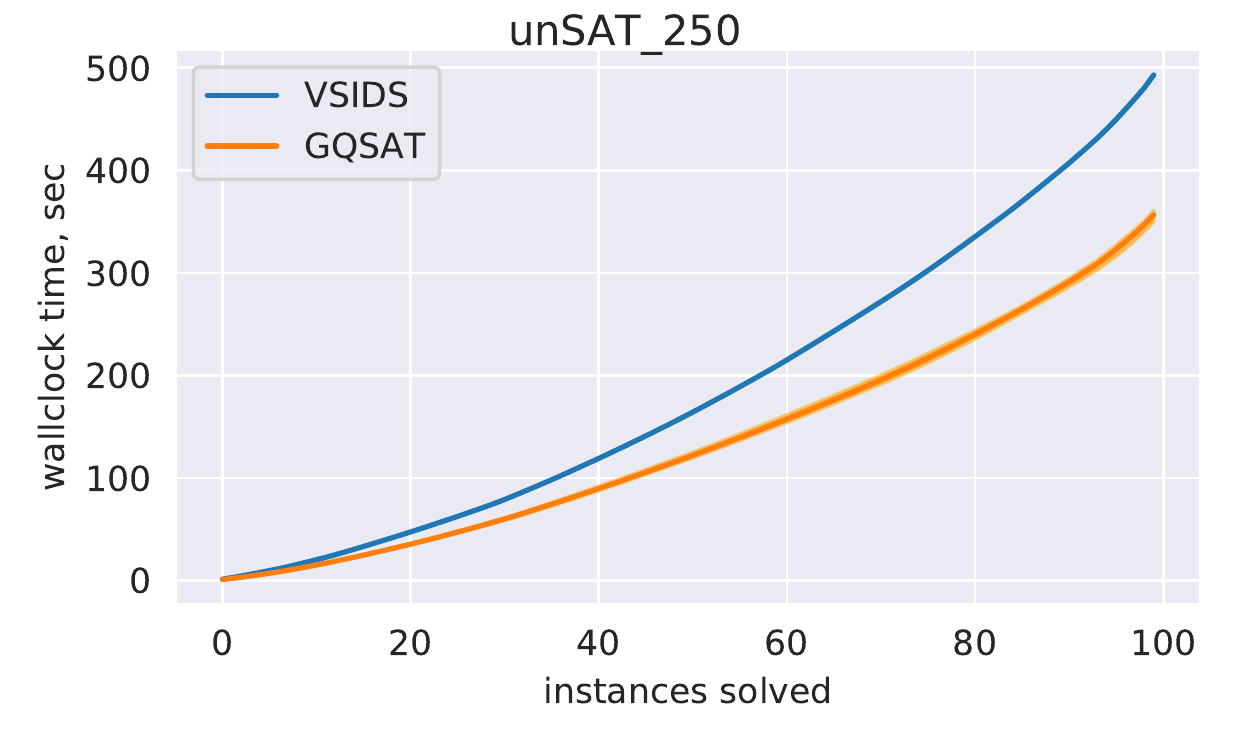}}
  \caption{\method{}'s \metric{} improvement (10 model calls) results in the wall clock time reduction. The curves show the averaged performance across five runs with the shade denoting the worst and the best runs.}
  \label{fig:cactus}
\end{figure}

Figure~\ref{fig:cactus} demonstrates that reduction in the number of iterations together with limiting the number of model calls results in wall clock time improvements for the datasets where the number of saved iterations is large enough to tolerate the network inference timings. 

\begin{figure}
\centering
\includegraphics[width=\columnwidth]{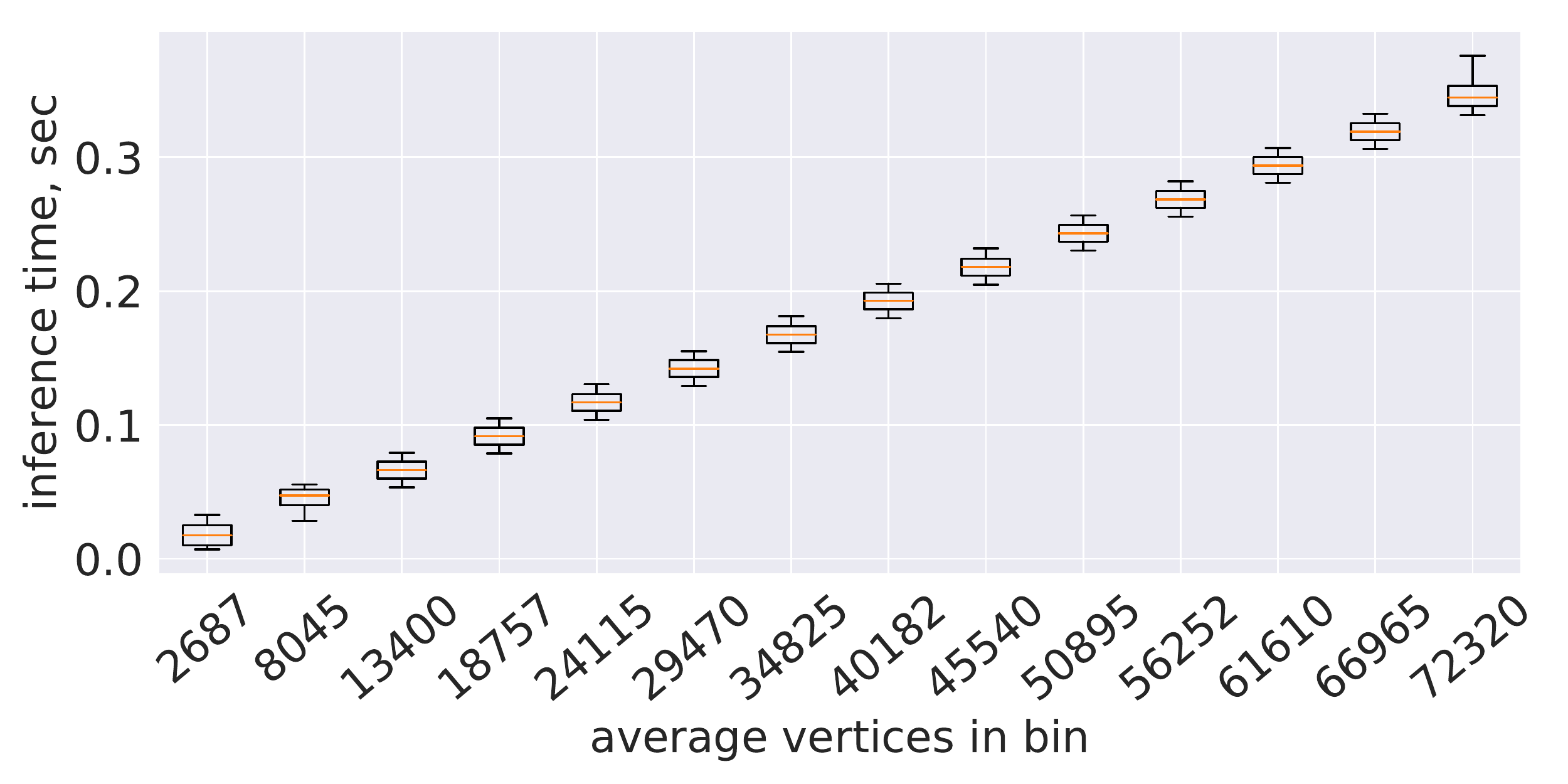}
\caption{\method{} inference time linearly depends on the number of vertices in the graph.}
\label{fig:timings}
\end{figure}

More generally, we can anticipate the settings in which \method{} will yield an improvement in wall clock time over VSIDS by analyzing the factors contributing to their runtime performance.
Assuming VSIDS's computational cost is negligible, we can compute the wall clock time of \minisat{} with VSIDS as follows: $W_{VSIDS} = \sum_{t=1}^{T}P(t)$, where $P(t)$ is the unit propagation time (a procedure of formula simplification after the branching decision is made).
Similarly, \method{} saves some fraction of iterations at the cost of added neural network inference time: $W_{\text{\method{}}} = \sum_{t=1}^{T/S}P(t) + \sum_{t=1}^{K}I(t)$,
where $S$ is the reduction of the number of iterations, $K$ is the number of our model forward passes ($K<<T$ for larger problems), and $I(t)$ is the network inference time.
Thus, \method{} leads to wall clock speed ups when the total inference time stays below the time spent on propagation for the reduced number of VSIDS decisions.
This seems plausible assuming that $T$'s growth is unbounded, $K<<T$ and linear dependence of $I(t)$ on the number of vertices.
To check the linear dependence, we generated $10^5$ graphs with characteristics similar to random 3-SAT problems (bipartite graph, each variable is connected to 13 clauses, and clause/variable ratio is 4).
Figure~\ref{fig:timings} confirms that the dependence is linear.

\section{Propagations per step}
\label{sec:propagations}
\begin{figure}
\centering
\subfigure[SAT 50-218]{
    \centering
    \includegraphics[width=0.54\columnwidth]{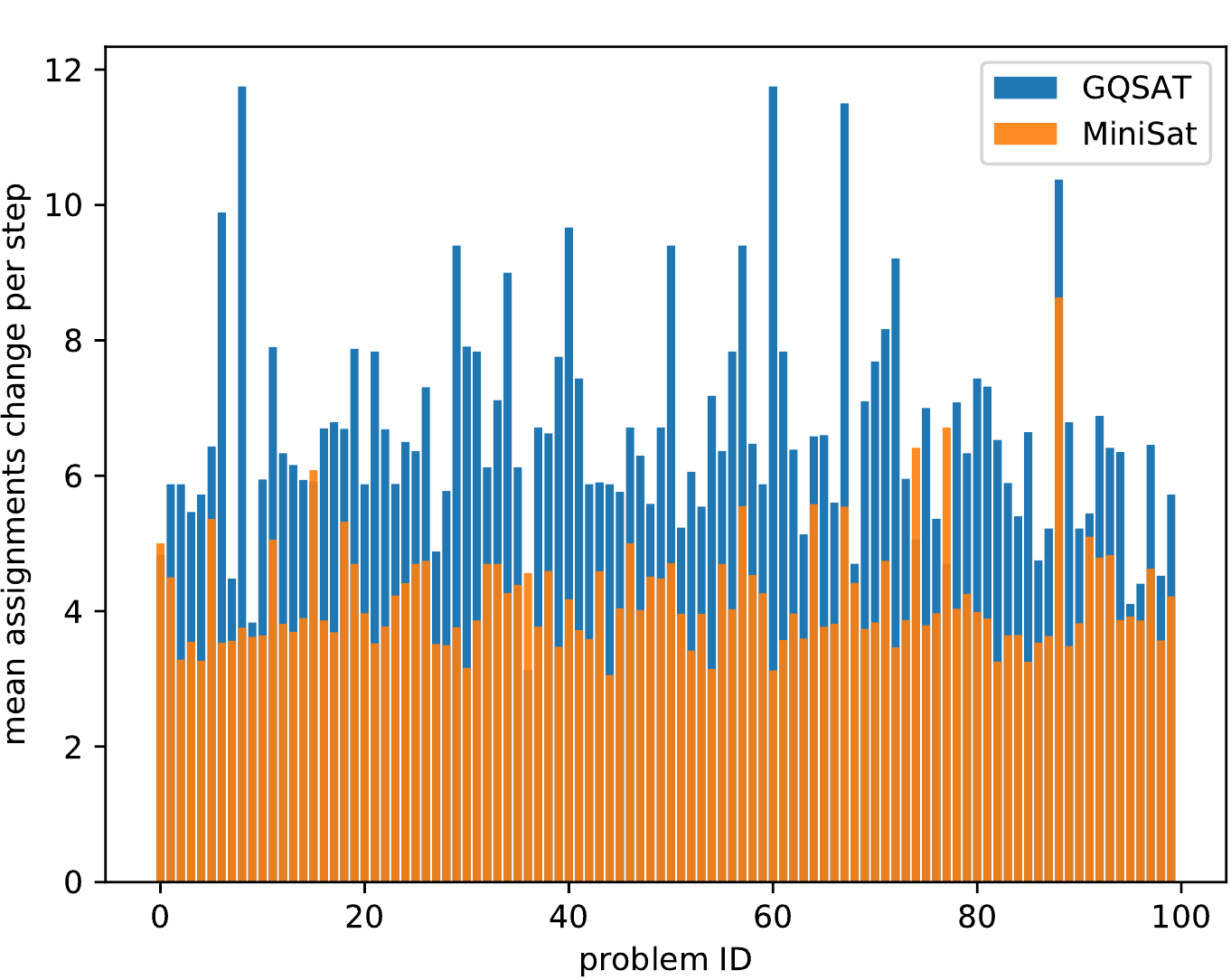}}%
\subfigure[unSAT 50-218]{
    \centering
    \includegraphics[width=0.54\columnwidth]{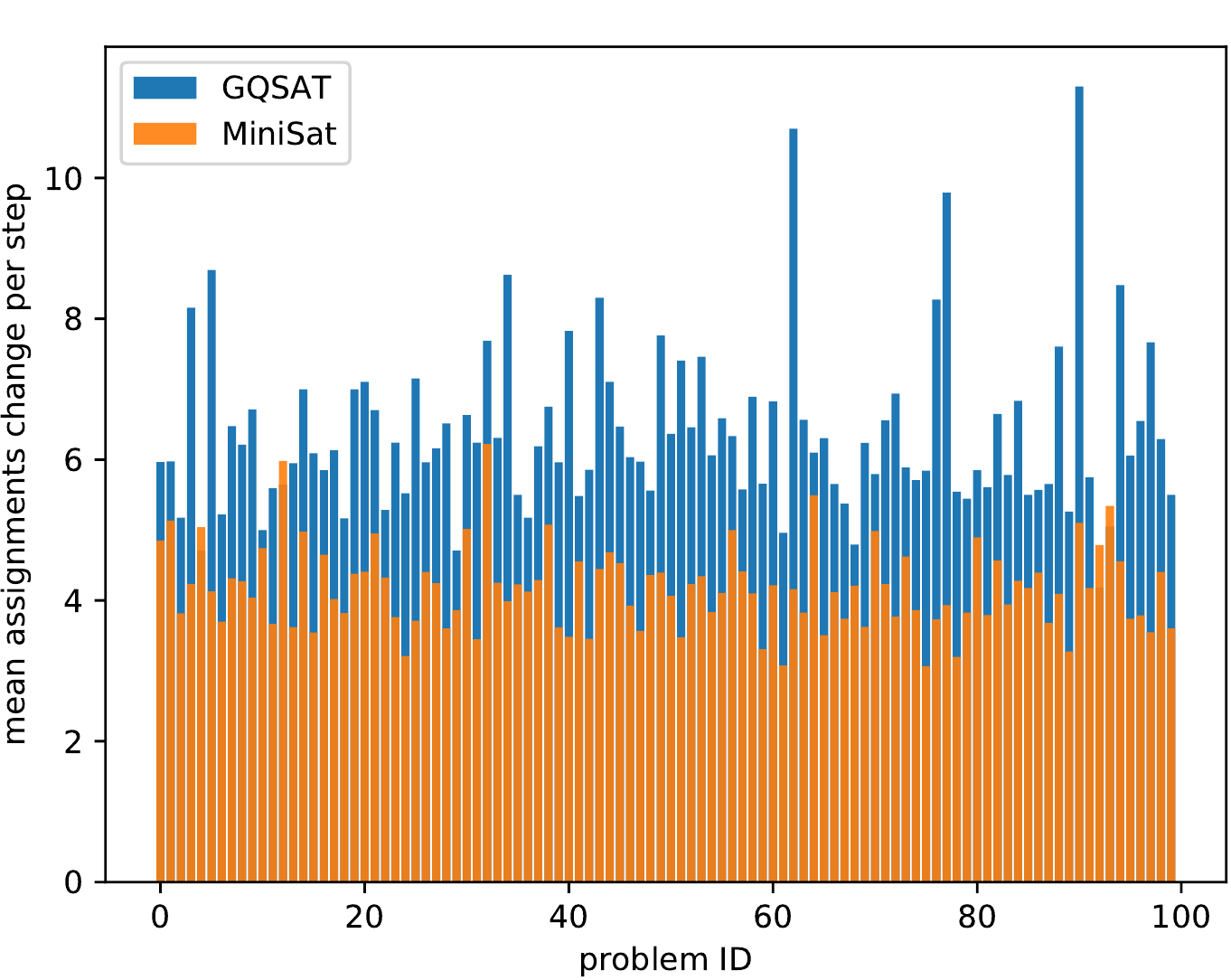}}
\subfigure[SAT 100-430]{
    \centering
    \includegraphics[width=0.54\columnwidth]{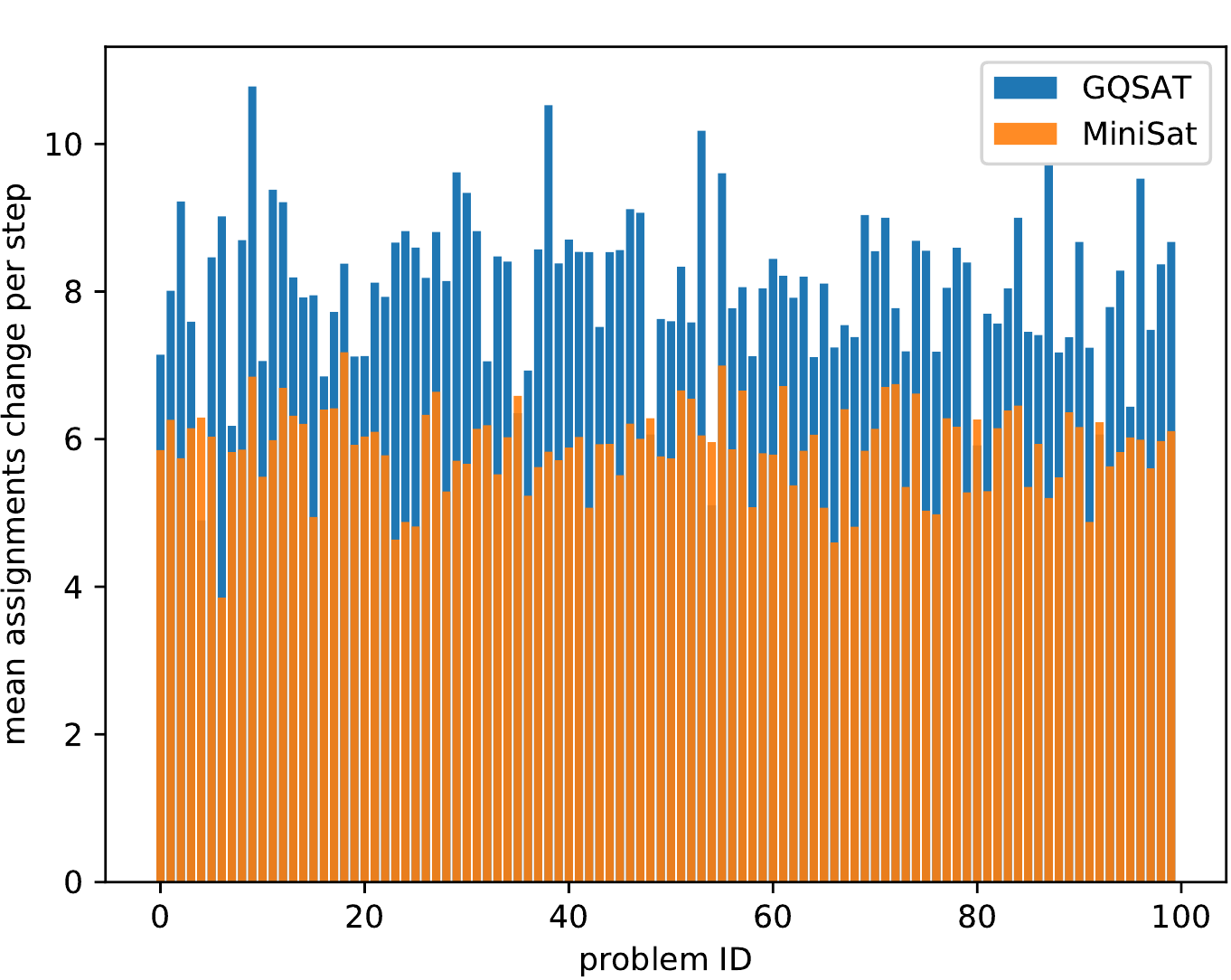}
}%
\subfigure[SAT 100-430]{
    \centering
    \includegraphics[width=0.54\columnwidth]{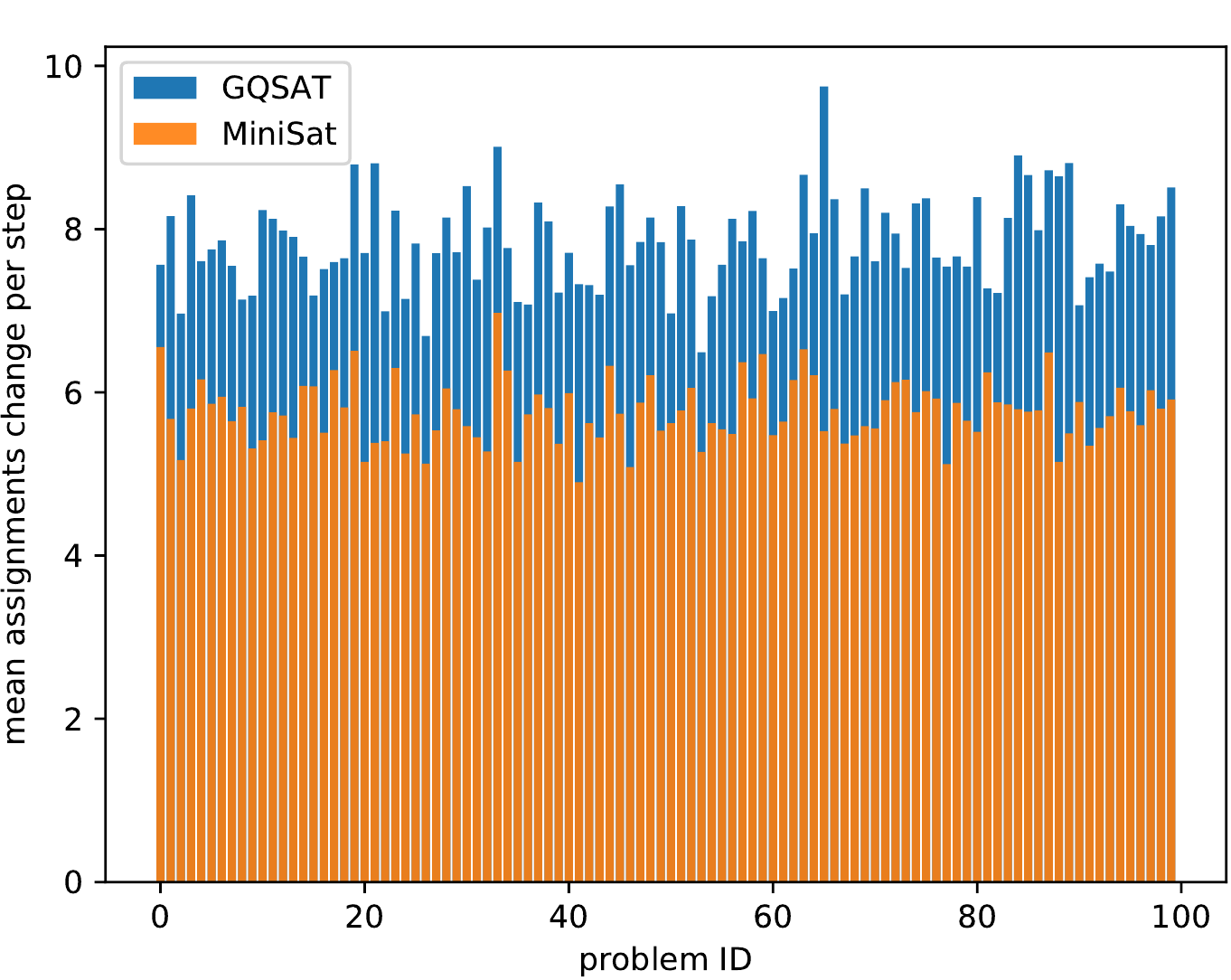}
}
\caption{Average number of variable assignments change per step for (un)SAT-50-218 and (un)SAT-100-430.}
\label{fig:props-per-step}
\end{figure}

Figure~\ref{fig:props-per-step} shows that on average using \method{} leads to more propagations per step than VSIDS.

\section{Reproducibility}
\label{sec:reproducibility}

\subsection{Model architecture}
\label{sec:arch}
\begin{figure}[h]
    \centering
    \scalebox{.8}{
    \begin{tikzpicture}     
      \Vertex[x=0, y=1, label=Core, fontscale=2, size=2, shape=rectangle, color=orange, shape=rectangle]{c}
      \Vertex[x=-3, y=-1.0, label=Encoder, fontscale=2, size=2, shape=rectangle, color=orange, shape=rectangle]{e}
      \Vertex[x=3, y=-1.0, label=Decoder, fontscale=2, size=2, shape=rectangle, color=orange, shape=rectangle]{d}

      \Vertex[x=-3, y=1, label=concat, fontscale=1.2, size=1, color=orange]{concat}

      \Edge[Direct, bend=0](e)(concat)
      \Edge[Direct, bend=0](concat)(c)
      \Edge[Direct, bend=40, label=$G^T$](c)(d)
      \Edge[Direct, bend=-100, label=$G^k$](c)(concat)
      
      \Vertex[x=-3, y=-3, label=$G$, Pseudo=true, fontscale=2, size=1]{g}
      \Edge[Direct, bend=0](g)(e)
      \Vertex[x=3, y=-3, label=$G'$, Pseudo=true, fontscale=2, size=1]{gp}
      \Edge[Direct, bend=0](d)(gp)
    \end{tikzpicture}
    }
    \caption{Encode-Process-Decode architecture. Encoder and Decoder are independent graph networks, i.e. MLPs taking whole vertex/edge data array as a batch. $k$ is the index of a message passing iteration. When concatenating for the first time, encoder output is concatenated with zeros.}
  \end{figure}
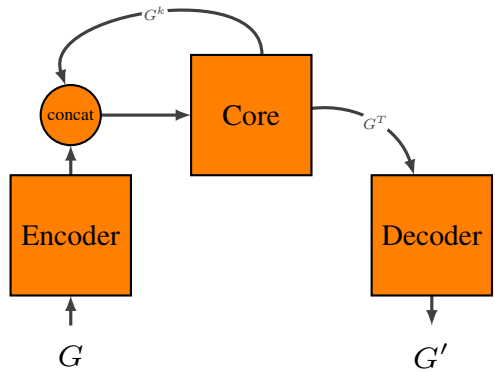

We use Encoder-Process-Decode architecture. 
Encoder and decoder are independent graph networks, i.e. MLPs taking whole vertex or edge feature matrix as a batch without message passing.
We call the middle part 'the core'.
The output of the core is concatenated with the output of the encoder and gets fed to the core again.
We describe all the hyperparameters in Appendix~\ref{sec:hyperparameters}.
We also plan to release the experimental code and the modified version of \minisat{} to use as a gym environment.

\subsection{Dataset}
\label{sec:dataset}

We split SAT-50-218 into three subsets: 800 training problems, 100 validation and 100 test problems.
For generalization experiments, we use 100 problems from all the other benchmarks.

For graph colouring experiments, we train our models using all problems from flat-75-180 dataset.
We select a model, given the performance on all 100 problems from flat-100-239.
So, evaluation on these two datasets should not be used to judge the performance of the method, and they are shown separately in Table~\ref{tab:structured}. 
All the data from the second part of the table was not seen by the model during training (flat-30-60, flat-50-115, flat-125-301, flat-150-360, flat-175-417, flat-200-479).

\subsection{Hyperparameters}
\label{sec:hyperparameters}

\begin{table*}
\centering
\caption{\method{} hyperparameters.}
\label{tab:hyperparameters}
\begin{tabular}{lll}
\textbf{Hyperparameter}&\textbf{Value}&\textbf{Comment}\\
\midrule
\textit{DQN}&&\\
\midrule
-- Batch updates & 50 000&\\
-- Learning rate & 0.00002&\\
-- Batch size & 64&\\
-- Memory replay size & 20 000&\\
-- Initial exploration $\epsilon$ & 1.0&\\
-- Final exploration $\epsilon$ & 0.01&\\
-- Exploration decay& 30 000& Environment steps.\\
-- Initial exploration steps & 5000 &Environment steps, filling the buffer, no training.\\
-- Discounting $\gamma$ & 0.99\\
-- Update frequency & 4 & Every 4th environment step.\\
-- Target update frequency & 10 & \\
-- Max decisions allowed for training & 500 & Used a safety against being stuck at the episode.\\
-- Max decisions allowed for testing & 500 & \makecell[tl]{Varied among [0, 10, 50, 100, 300, 500, 1000] \\for the experiment on Figure~\ref{fig:k_steps}.} \\
-- Step penalty size $p$ & -0.1&\\
\midrule
\textit{Optimization}&&\\
\midrule
-- Optimizer & Adam & \\
-- Adam betas & 0.9, 0.999&Pytorch default.\\
-- Adam eps&1e-08&Pytorch default.\\
-- Gradient clipping & 1.0& 0.1 for training on the graph coloring dataset.\\
-- Gradient clipping norm & $L_2$&\\
-- Evaluation frequency&1000\\
\midrule
\textit{Graph Network}&&\\
\midrule
-- Message passing iterations&4&\\
-- Number of hidden layers for GN core &1&\\
-- Number of units in GN core &64&\\
-- Encoder output dimensions &32& For vertex, edge and global updater.\\
-- Core output dimensions & 64,64,32 & For vertex, edge and global respectively.\\
-- Decoder output dimensions &32&\makecell[tl]{For vertex updater, since only Q values are used,\\ no need for edge/global updater.}\\
-- Activation function&ReLU&For everything but the output transformation.\\
-- Edge to vertex aggregator $\rho_{e \rightarrow v}$ & sum &\\
-- Variable to global aggregator $\rho_{v \rightarrow u}$ & average &\\
-- Edge to global aggregator $\rho_{e \rightarrow u}$ & average &\\
-- Normalization & Layer Normalization& After each GN updater
\end{tabular}
\end{table*}

Table~\ref{tab:hyperparameters} contains all the hyperparameters necessary to replicate our results.

\subsection{\method{} pseudocode}
\begin{algorithm}[]                   
   \caption{\method{} Action Selection}                        
   \label{alg:inference}                                 
\begin{algorithmic}
    \STATE {\bfseries Input:} graph network $GN_{\theta}$, state graph $G_s := (V,E,U)$, \\ with vertex features $V = [V_{vars}, V_{clauses}]$, edge features $E$, and a global compontent $U$.
    \\\hrulefill
    \STATE $V',E',U' = GN(V,E,U)$;
    \STATE $VarIndex,VarPolarity = \arg\max_{ij}V_{vars}'$;
    \STATE \textbf{Return} $VarIndex,VarPolarity$;
\end{algorithmic}                       
\end{algorithm}

\begin{algorithm}
   \caption{\method{} Training Procedure}                        
   \label{alg:training}                                 
\begin{algorithmic}
    \STATE {\bfseries Input:} Set of tasks $\mathcal{S}\sim\mathcal{D}(\phi, (un)SAT,n_{vars},$$ n_{clauses})$ split into $\{\mathcal{S}_{train}, \mathcal{S}_{validation}, \mathcal{S}_{test}\}$, $\phi$ is the task family (e.g. random 3-SAT, graph coloring). All hyperparameters are from Table~\ref{tab:hyperparameters}. 
    \STATE Randomly Initialize Q-network $GN_{\theta}$;
    \STATE $updates = 0$;
    \STATE $totalEnvSteps=0$;
    \REPEAT
        \REPEAT
            \STATE Sample a SAT problem $p \sim \mathcal{S}_{train}$;
            \STATE Initialize the environment $env = SatEnv(p)$;
            \STATE Reset the environment $s = env.reset()$;
            \STATE take action \\$a = \begin{cases}
                    random(\mathcal{A}), \text{with probability}~\epsilon \\
                    selectAction(s), \text{with probability}~1-\epsilon \\
                    \end{cases}$
            \STATE Take env step $s', r, done$ = $env.step(a)$;
            \STATE $totalEnvSteps+=1$;
            \STATE dump experience $buffer.add(s, s', r, done, a)$;
            \IF{$totalEnvSteps$ mod $updateFreq == 0$;}
                \STATE Do a DQN update;
            \ENDIF 
            \IF{$totalEnvSteps$ mod $validateFreq == 0$;}
                \STATE Evaluate $GN_{\theta}$ on $\mathcal{S}_{validation}$;
            \ENDIF
        \UNTIL{Proved SAT/unSAT ($done$ is \textit{True})}
    \UNTIL{$updates == totalBatchUpdates$}
    \STATE Pick the best model $GN_{\theta}$ given validation scores;
    \STATE Test the model $GN_{\theta}$ on $\mathcal{S}_{test}$;
\end{algorithmic}                       
\end{algorithm} 

% \clearpage
% \bibliographystyle{abbrvnat}
% \bibliography{neurips_2020}